\documentclass[letterpaper]{article} 
\usepackage{aaai2026}  
\usepackage{times}  
\usepackage{helvet}  
\usepackage{courier}  
\usepackage[hyphens]{url}  
\usepackage{graphicx} 
\usepackage{natbib}  
\usepackage{caption} 
\usepackage{algorithm}
\usepackage{algorithmic}

\usepackage{newfloat}
\usepackage{listings}
\DeclareCaptionStyle{ruled}{labelfont=normalfont,labelsep=colon,strut=off} 
\floatstyle{ruled}
\newfloat{listing}{tb}{lst}{}
\floatname{listing}{Listing}
\usepackage{amsmath}
\usepackage{amsfonts}
\usepackage{booktabs}

\title{\model{}: Hydrological Domain-Conditioned Modulation for Cross-Reservoir Inflow Prediction}
\author {
    Pengfei Hu\textsuperscript{\rm 1,\rm 2}\thanks{Work done while interning at Oak Ridge National Laboratory},
    Fan Ming\textsuperscript{\rm 1},
    Xiaoxue Han\textsuperscript{\rm 2},
    Chang Lu\textsuperscript{\rm 2},
    Yue Ning\textsuperscript{\rm 2},
    Dan Lu\textsuperscript{\rm 1}
}
\affiliations {
    \textsuperscript{\rm 1}Oak Ridge National Laboratory, Oak Ridge, TN, USA\\
    \textsuperscript{\rm 2}Stevens Institute of Technology, Hoboken, NJ, USA\\
    \{phu9, xhan26, clu13, yue.ning\}@stevens.edu, \{fanm, lud1\}@ornl.gov
}

\newcommand{\model}{\texttt{HydroDCM}}

\begin{document}

\maketitle

\begin{abstract}
Deep learning models have shown promise in reservoir inflow prediction, yet their performance often deteriorates when applied to different reservoirs due to distributional differences, referred to as the domain shift problem. 
Domain generalization (DG) solutions aim to address this issue by extracting domain-invariant representations that mitigate errors in unseen domains.
However, in hydrological settings, each reservoir exhibits unique inflow patterns, while some metadata beyond observations like spatial information exerts indirect but significant influence.
This mismatch limits the applicability of conventional DG techniques to many-domain hydrological systems. 
To overcome these challenges, we propose \model{}, a scalable DG framework for cross-reservoir inflow forecasting. 
Spatial metadata of reservoirs is used to construct pseudo-domain labels that guide adversarial learning of invariant temporal features. 
During inference, \model{} adapts these features through light-weight conditioning layers informed by the target reservoir’s metadata, reconciling DG’s invariance with location-specific adaptation.
Experiment results on 30 real-world reservoirs in the Upper Colorado River Basin demonstrate that our method substantially outperforms state-of-the-art DG baselines under many-domain conditions and remains computationally efficient.
\end{abstract}

\begin{links}
    \link{Code}{https://github.com/humphreyhuu/HydroDCM}
\end{links}

\section{Introduction}
Reservoir inflow prediction is a critical task in water resources management, directly influencing flood control, water allocation, and hydropower generation~\citep{gupta2022two, fan2022identifying, latif2023review}. 
Reliable forecasts help mitigate the risks of extreme events and improve overall water-use efficiency~\citep{fan2023explainable}.
Even though deep learning predictive methods, such as Long Short-Term Memory (LSTM) networks, temporal convolutional architectures, and Transformers, have shown strong performance on single reservoirs, their effectiveness is compromised in data-sparse environments, particularly for reservoirs with limited historical records or ungaged systems with no observational data~\citep{fan2023advancing, fan2025enhancing, li2024data}. 

In recent years, research interest has increasingly focused on multi-reservoir inflow prediction~\citep{kratzert2018rainfall, hess-28-4187-2024}. 
For instance, graph neural networks are adopted to enable spatial information sharing among reservoirs in hydrological settings~\citep{sun2021explore}. 
However, when applied to previously ``unseen'' reservoirs, the learned inter-reservoir edges often lead to performance degradation, and even fail to offer reliable forecasts \citep{kratzert2019toward}.

This challenge is also known as the domain shift problem, where reservoir-inflow patterns are usually decided by unique climatic conditions, geographic locations, catchment operations~\citep{khan2022impact, martyushev2024multiscale}. 
Consequently, a predictive model trained on one reservoir cannot be directly applied to another without addressing domain shifts. 
Moreover, target (training) reservoirs may lack sufficient historical data, requiring models to generalize observations from other source (test) reservoirs~\citep{latif2023review}. 
It aligns closely with the goal of domain generalization (DG) problems, which aims to enhance robustness of models and provide accurate predictions on unseen target (test ) data~\citep{zhou2022domain, zhao2024domain,hu2025udoncare}.
However, applying regular DG solutions may encounter two challenges in hydrology:

\begin{itemize}
    \item \textbf{Many-Domain Categorization.}
    Reservoir networks often contain many sites, sometimes reaching dozens or even more than one hundred reservoirs~\citep{fan2023explainable, sun2021explore}. 
    It is ideal to treat each reservoir as an independent domain to fully capture domain-specific variability. 
    Some DG methods like meta-learning techniques can accommodate this setting, but their computational overhead increases rapidly with the number of reservoirs, which makes them impractical for large-scale hydrological systems. 
    Other methods are developed under the assumption of a few latent domains, which makes model hard to remove domain covariates from input features. 
    \item \textbf{Reliance on Metadata.}
    Reservoirs also have abundant spatial and environmental metadata beyond observations, and such information provides guidance for models to distinguish reservoirs rather than relying solely on different hydrometeorological dynamics~\cite{tsai2014including, moradi2020long}. 
    However, existing DG methods primarily focus on invariant-feature learning and disregard auxiliary metadata that may capture essential hydrological differences. 
    They typically overlook the potential of different sources of information that could mitigate domain shift in reservoir inflow forecasting. 
\end{itemize}

Unlike conventional DG works that focus on a small number of domains, this work explores a new setting for learning generalizable models from a substantially larger number of domains across reservoirs, addressing the urgent need for hydrological operations on newly-built reservoirs. 
To better mitigate domain shifts in hydrological practices, we develop \model{}, a DG framework tailored for reservoir inflow prediction that can (i) handle the diversity of reservoir covariates efficiently, and (ii) integrate explicit domain features into the learning process without sacrificing domain-invariance in the latent space. 
Concretely, we distinguish latent domains for reservoirs by modeling geographic features, and learn (reservoir-invariant) label representations by following the adversarial training paradigm. 
Then, future inflow predictions are modulated by both scaling and shifting factors with spatial meta attributes. 
We summarize our main contributions below: 
\begin{enumerate}
    \item To the best of our knowledge, ours is the first work that extends domain generalization (DG) to hydrological applications. Our setting is challenging: We handle a broader range of domains by incorporating meta-attributes (e.g. geographical features) compared to most existing DG studies. 
    \item \model{} explicitly tackles this many-domain challenge through attribute injection, bridging the gap between standard DG approaches and the specific requirements of reservoir inflow forecasting. 
    \item We evaluate our method on real-world reservoir datasets in the Upper Colorado River Basin, where \model{} outperforms all DG baselines across three reservoirs using only three years of observations.
\end{enumerate}

\section{Related Work}
\subsection{Domain Generalization (DG)}
DG aims to learn representations that are invariant across domains so that the model can be extrapolated to domains that have not yet been explored. 
The rationale is that features invariant to the distribution shift among source domains should remain robust when facing any unseen target domain shift~\citep{zhou2022domain, zhao2024domain}.
Existing DG methods can be broadly categorized into three clusters: 
(1) Domain-invariant feature learning~\citep{ganin2016domain, arjovsky2019invariant, matsuura2020domain}, which minimizes inter-domain discrepancies through alignment objectives or adversarial learning; 
(2) Meta-learning methods~\citep{li2018learning, shu2021open, khoee2024domain}, which simulates domain shifts during training to enhance model robustness across unseen environments;
and (3) Data augmentation methods~\citep{volpi2018generalizing, li2021simple, su2023rethinking}, which increases domain diversity to promote invariance and robustness. 
Recent studies have extended DG to diverse fields, such as computer vision, fault diagnosis, and healthcare prediction \citep{zhao2024domain, hu2025udoncare}, demonstrating its ability to capture domain-invariant patterns under varying distributions.

Despite these advances, applying domain generalization problems to hydrological forecasting is far from trivial by dealing with a large number of domains by considering each reservoir as single domain~\citep{kratzert2019toward, li2022invariant}. 
This challenge motivates us to incorporate meta attributes as pseudo-domain identifiers to capture the domain information explicitly in our proposed method. 

\subsection{Reservoir Inflow Prediction}
Most works acknowledge that inflow patterns differ significantly across reservoirs because each site follows its own hydrological, climatic, and operational dynamics~\citep{gupta2022two, latif2023review, fan2023advancing}. 
Early deep learning techniques were typically trained and evaluated on individual reservoirs~\citep{kratzert2018rainfall, fan2025enhancing}. 
However, such training strategy highly relies on the quantity and quality of observations, suffering from significant performance degradation on reservoirs with few records. 
Motivated by this challenge, recent studies have explored multi-reservoir forecasting to improve model generalization and jointly capture shared inflow patterns. 
For example, graph neural networks~\citep{sun2021explore, hu2025adaptive} are used to propagate spatial dependencies in terms of neighboring features, while others incorporate geographical features, climate indices, and terrain characteristics to enhance hidden representation~\citep{wang2022unraveling, zeng2023evaluating}. 
These studies indicate that incorporating reservoir-specific attributes beyond training data is crucial for getting accurate and robust predictions.

However, most works focus on overall performances of the reservoir network, and fewer of them evaluate model generalization on data-scarce or unseen reservoirs. 
Our method bridges this gap by elevating spatial information from an auxiliary feature to a generalization-driving signal, enabling robust cross-reservoir inflow prediction even under distributional shifts and limited observations.

\section{Preliminary}
\begin{figure*}[t]
    \centering
    \includegraphics[width=\linewidth]{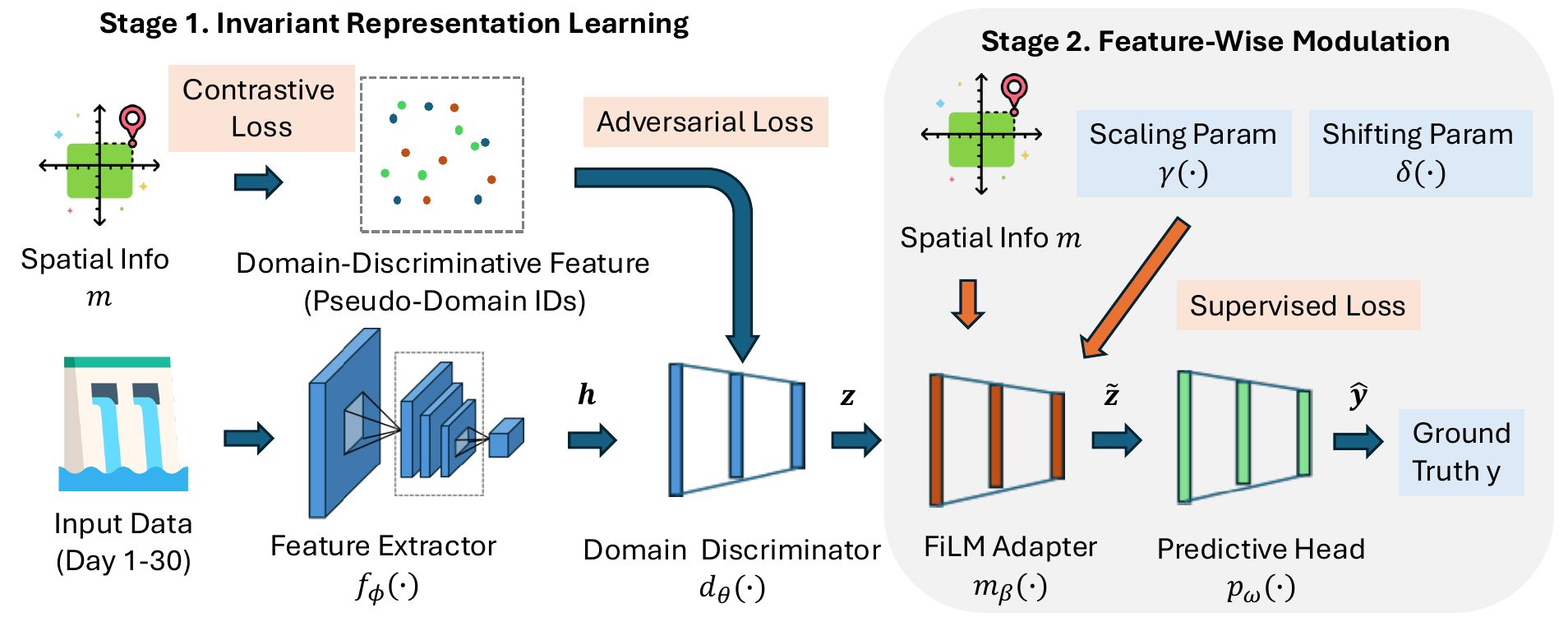}
    \caption{Overview of \model{} architecture for reservoir inflow prediction. Given the observations $\mathbf{X}_i$, (1) the feature encoder $f_\phi(\cdot)$ extracts temporal representations $\mathbf{h}_i$, which are refined into invariant features $\mathbf{z}_i$ through adversarial learning with $d_\theta(\cdot)$ using spatial metadata $\mathbf{s}_i$ as pseudo-domain labels; (2) modulation parameters $\{\gamma(\mathbf{s}_i), \delta(\mathbf{s}_i)\}$ in the FiLM adapter $m_\beta(\cdot)$ adjust $\mathbf{z}_i$ into $\tilde{\mathbf{z}}_i = \gamma(\mathbf{s}_i)\odot \mathbf{z}_i + \delta(\mathbf{s}_i)$, which is then passed to the predictive head $p_\omega(\cdot)$ for estimating future inflow $\hat{y}_i$.
}
    \label{fig:framework}
\end{figure*}

\subsubsection{Inflow Observation \& Task Definition.}
Following regular reservoir inflow forecasting settings~\cite{fan2023advancing, fan2023explainable}, we construct a dataset $\mathcal{X} = \{\mathbf{X}_1, \mathbf{X}_2, \dots, \mathbf{X}_N\}$ containing $N$ reservoirs in certain river basin. 
For the $i$-th reservoir, the historical record is denoted as $\mathbf{X}_i = \{\mathbf{x}_{i,t-T+1}, \dots, \mathbf{x}_{i,t}\} \in \mathbb{R}^{T \times F}$, where $F$ represents the number of hydrometeorological features (e.g., precipitation, temperature, and past inflows) over a rolling window of $T=30$ days. 
Note that observation of single reservoir $\mathbf{X}_i$ should be normalized prior to the model training phase.
These temporal sequences serve as model inputs for forecasting the future inflow. 
The prediction targets are the daily inflow values $\mathbf{y}_{i,t+k}$ for the next $H=7$ days, where $k \in \{1,\dots,H\}$ indicates the forecasting horizon. 
During training, models can access a training (source) subset within $|\mathcal{S}|$ reservoirs in total, while we evaluate generalization on the test (target) subset of $|\mathcal{T}|$ reservoirs.
Each reservoir is treated as a distinct domain $\mathcal{D}_i \in \mathcal{S} \cup \mathcal{T}$ because inflow patterns are governed by unique climatic, geographic, and operational factors. 

\subsubsection{Meta Attributes.}
Besides temporal data, each reservoir is also associated with a spatial attribute vector 
$\mathbf{s}_i \in \mathbb{R}^M$, which describes domain knowledge independent of the time series observations. 
In this paper, we consider geographical coordinates and elevation as meta attributes, but other environmental descriptors (e.g. basin characteristics, mean annual precipitation) could be also involved by our method.

\subsubsection{Problem Formulation of DG.}
Similar to other DG frameworks~\cite{matsuura2020domain, hu2025udoncare}, a model $f_\phi(\cdot)$ parameterized by $\phi$ is trained on source data drawn from the distribution $P_{tr}$ to minimize the empirical loss:
\begin{equation}
    \arg \min_{\phi} \ \mathbb{E}_{(x, y) \sim P_{tr}} \bigl[ \ell(f_\phi(\mathbf{x}), y) \bigr],
    \label{eq:dg_arg}
\end{equation}
where $\ell(\cdot)$ denotes a supervised loss function (e.g. mean squared error). 
The final objective is to ensure that the trained model generalizes to unseen test data sampled from another distribution $P_{te}$. 
However, in hydrological forecasting, the domain shift is typically pronounced because reservoir inflows are influenced by diverse geographic conditions and climatic variations, resulting in $P_{tr} \neq P_{te}$. 
Therefore, the DG objective becomes learning invariant (label) representations 
$\mathbf{z}$ from hidden features $\mathbf{h} = f_\phi(\mathbf{x})$, 
while filtering out spurious, domain-specific features (covariates) that hinder cross-reservoir generalization.

\section{Methodology}
\model{} consists of four modules: a temporal feature extractor $f_\phi(\cdot)$, a domain discriminator $d_\theta(\cdot)$, a feature-wise linear modulation (FiLM) adapter $m_\beta(\cdot)$, and a predictive head $p_\omega(\cdot)$.
Our method operates through two general steps: 
\begin{itemize}
    \item \textit{Step 1. Adversarial-Training Generalization.} The discriminator $d_\theta(\cdot)$ guides the encoder $f_\phi(\cdot)$ to remove domain-related information from latent representations, forcing the extracted features $\mathbf{z}$ to focus on label-relevant hydrological dynamics. 
    \item \textit{Step 2. Domain-Conditioned Modulation.} The adapter $m_\beta(\cdot)$ introduces lightweight, spatially-aware modulation at inference time, adapting the invariant representation $\mathbf{z}$ to the target reservoir’s specific geographic and climatic context for prediction through $p_\omega(\cdot)$.
\end{itemize}
During training, models access a source subset of reservoirs with adequate records, while others with limited data (target domains) are used for evaluation only. 
It ensures robust forecasting performance on unseen reservoirs on top of existing works. 
The overall architecture is illustrated in Figure~\ref{fig:framework}.

\subsection{Temporal Extraction}
Given the historical observation matrix $\mathbf{X}_i$ for the $i$-th reservoir, the temporal encoder first maps the input sequence into a compact hidden representation
\begin{equation}
    \mathbf{h}_i = f_\phi(\mathbf{X}_i),
\end{equation}
where $\mathbf{X}_i \in \mathbb{R}^{T \times F}$ contains $F$ hydrometeorological features over a rolling window of $T$ days. 
Note that \model{} can serve as a plug-and-play module for most hydrological models and can be extended to additional tasks, since $f_\phi(\cdot)$ can adopt different architectures, ensuring flexibility with existing forecasting frameworks. 
The resulting hidden state $\mathbf{h}$ preserves long-term temporal dependencies while abstracting away short-term noise, preparing the representations for domain-invariant learning in the next step.

\subsection{Step 1: Adversarial-Training Generalization}
The first stage aims to extract representations that capture essential hydrological dynamics while filtering out domain-specific noise. 
Directly assigning a unique domain label to each reservoir is computationally expensive and may lead to overfitting when the number of reservoirs is large. 
To address this, we represent each reservoir using its spatial metadata $\mathbf{s}$ (e.g., longitude, latitude, and elevation), which acts as a pseudo-domain identifier. 
The metadata vector $\mathbf{s}$ is projected into a discriminative embedding space as
\begin{equation}
  \mathbf{v}_i = \mathbf{W}\;[\mathbf{s}_i:\mathbf{X}_i]\;+ \mathbf{b},  
\end{equation}
where $\mathbf{W}, \;\mathbf{b}$ denote weights and bias in a linear classifier, and $[\cdot:\cdot]$ denotes the dimension-wise concatenation.
$\mathbf{s}$ is then optimized via a contrastive objective to ensure that reservoirs with similar geographical characteristics are positioned closer in this latent space. 
The corresponding contrastive loss is defined as:
\begin{align}
L_{\text{con}} = - \sum_{i=1}^{N} 
\log \frac{
\exp(\text{sim}(\mathbf{v}_i, \mathbf{v}_i^+)/\tau)
}{
\exp(\text{sim}(\mathbf{v}_i, \mathbf{v}_i^+)/\tau) 
+ \sum_{j\in\mathcal{N}}
},
\end{align}
where $\sum_{j\in\mathcal{N}}$ equals $\sum_{j=1}^{M}\exp(\text{sim}(\mathbf{v}_i, \mathbf{v}_j^-)/\tau)$ within the negative sample set $\mathcal{N}$, $\text{sim}(\cdot,\cdot)$ denotes cosine similarity, and $\tau$ is the temperature coefficient controlling contrastive sharpness. 
Positive and negative pairs $(\mathbf{v}_i^+, \mathbf{v}_j^-)$ correspond to same-domain and different-domain reservoirs, respectively.
Furthermore, adversarial learning encourages the encoder to suppress domain cues in $\mathbf{h}$. 
The discriminator $d_\theta(\cdot)$ is trained to misclassify the domain of the $i$-th reservoir based on soft identifiers $\mathbf{v}_i$, as shown in equation~\eqref{eq:adv_loss}:
\begin{equation}
L_{\text{adv}} 
= -\mathbb{E}_{(\mathbf{x}_i, \mathbf{v})} 
\|\, d_\theta(\mathbf{h}_i) - \mathbf{v}_i \,\|_2^2,
\label{eq:adv_loss}
\end{equation}
where $\mathbf{v}_i$ represents the soft pseudo-domain labels projected from the spatial metadata and $\|\cdot \|_2^2$ denotes the Euclidean Norm. 
During optimization, gradients are propagated only through the discriminator parameters $\theta$, which are updated in the reverse direction of the prediction objective. 
This training encourages $d_\theta(\cdot)$ to fail at distinguishing pseudo-domains, thereby preserving label hydrological signals. 
Through the interplay between $L_{\text{con}}$ and $L_{\text{adv}}$, the resulting feature $\mathbf{z}$ captures temporal dependencies essential for inflow prediction while filtering out spurious, reservoir-specific information. 
Therefore, such adversarial–contrastive formulation strikes a balance between domain invariance and spatial discriminability.

\subsection{Step 2: Domain-Conditioned Modulation}
While the previous step enforces invariance, the resulting features $\mathbf{z}$ may underfit domain-specific nuances that are hydrologically meaningful. 
To restore such information in a controlled manner, Stage~2 introduces domain-conditioned modulation using spatial metadata as external guidance. 
Specifically, the FiLM adapter $m_\delta(\cdot)$ generates scaling and shifting coefficients $m_\delta(\mathbf{s}_i):= \{\gamma(\mathbf{s}_i), \, \delta(\mathbf{s}_i)\}$ conditioned on the reservoir’s attributes. 
These two factors modulate the invariant representation with few parameters through a feature-wise affine transformation
\begin{equation}
    \tilde{\mathbf{z}}_i = \gamma(\mathbf{s}_i) \odot \mathbf{z}_i + \delta(\mathbf{s}_i),
\end{equation}
where $\odot$ denotes element-wise multiplication. 
This operation enables the model to incorporate spatial priors—such as topographic gradients and climatic variability—into the prediction space without reintroducing domain bias.
The modulated features are then fed into the predictive head $p_\omega(\cdot)$ to estimate the future inflow
\begin{equation}
\hat{y}_i = \text{MLP}(\tilde{\mathbf{z}}_i) = p_\omega(\tilde{\mathbf{z}}_i),
\end{equation}
where $\text{MLP}(\cdot)$ stands for multi-layer perceptron.
The corresponding supervised regression loss is given by:
\begin{align}
L_{\text{sup}} = \mathbb{E}_{(\mathbf{x}_i, y)}\big[\ell\big(p_\omega(\tilde{\mathbf{z}_i}), y_i\big)\big] 
= \frac{1}{N} \sum_{i=1}^{N} (y_i - \hat{y}_i)^2,
\end{align}
where $\ell(\cdot,\cdot)$ denotes mean squared error (MSE). 
By parameterizing $\gamma(\mathbf{s})$ and $\delta(\mathbf{s})$, the model adaptively captures how spatial context modulates hydrological responses, achieving improved robustness on unseen reservoirs with limited observations.
This step thus complements the pitfalls of invariant features by injecting domain-aware flexibility without reintroducing spurious correlations.

\subsection{Training \& Inference}  
The overall optimization integrates three complementary objectives: contrastive separation, adversarial alignment, and supervised prediction. Each objective plays a distinct role in enhancing model generalization.
To train models considering both accuracy and model generalization comprehensively, the complete loss function is 
\begin{align}
L_{\text{total}} = 
\lambda_{\text{con}} L_{\text{con}} 
+ \lambda_{\text{adv}} L_{\text{adv}} 
+ \lambda_{\text{sup}} L_{\text{sup}},
\end{align}
where $\lambda_{\text{con}}$, $\lambda_{\text{adv}}$, and $\lambda_{\text{sup}}$ are scalar weights controlling the trade-off among these objectives. 

During training, all modules ($f_\phi$, $d_\theta$, $m_\delta$, and $p_\omega$) are jointly optimized in an end-to-end manner. 
The adversarial and contrastive losses regularize the feature extractor to produce invariant yet discriminative embeddings, while the supervised term ensures predictive fidelity. 
At inference time, the domain discriminator is discarded, and only $f_\phi(\cdot)$, $m_\delta(\cdot)$, and $p_\omega(\cdot)$ are retained. 
This design ensures computational efficiency while maintaining the learned balance between invariance and domain adaptation. 
Consequently, \model{} generates hydrologically consistent forecasts for unseen reservoirs, even in scenarios of sparse or non-overlapping observations.

\section{Experiment Results}
\subsection{Dataset}
In this paper, we focus on reservoirs in the Upper Colorado River Basin, which spans Colorado, New Mexico, Utah, and Wyoming, supplying fresh water to nearly 40 million people for hydropower generation, flood control, and irrigation. 
Reservoirs are selected based on complete hydrological records of daily inflow data with minimal gaps (fewer than ten days) obtained from the U.S. Bureau of Reclamation water operation archive. 
We collect precipitation and temperature from the AN81d dataset of the PRISM model, which provides 4 km resolution coverage across the basin~\cite{daly2013prism}.
The basin includes: Big Sandy Reservoir (BSR), Causey Reservoir (CAU), Crystal Reservoir (CRY), Deer Creek Reservoir (DCR), Dillon Reservoir (DIL), Echo Reservoir (ECH), East Canyon Reservoir (ECR), Flaming Gorge Reservoir (FGR), Fontenelle Reservoir (FON), Green Mountain Reservoir (GMR), Hyrum Reservoir (HYR), Jordanelle Reservoir (JOR), Joes Valley Reservoir (JVR), Lost Creek Reservoir (LCR), Lemon Reservoir (LEM), McPhee Reservoir (MCP), Meeks Cabin Reservoir (MCR), Navajo Reservoir (NAV), Pineview Reservoir (PIN), Red Fleet Reservoir (RFR), Ridgway Reservoir (RID), Rockport Reservoir (ROC), Ruedi Reservoir (RUE), Scofield Reservoir (SCO), Silver Jack Reservoir (SJR), Starvation Reservoir (STA), Steinaker Reservoir (STE), Taylor Park Reservoir (TPR), Upper Stillwater Reservoir (USR), and Vallecito Reservoir (VAL).

\subsubsection{Data Split.}
\begin{figure}[t]
    \centering
    \includegraphics[width=\linewidth]{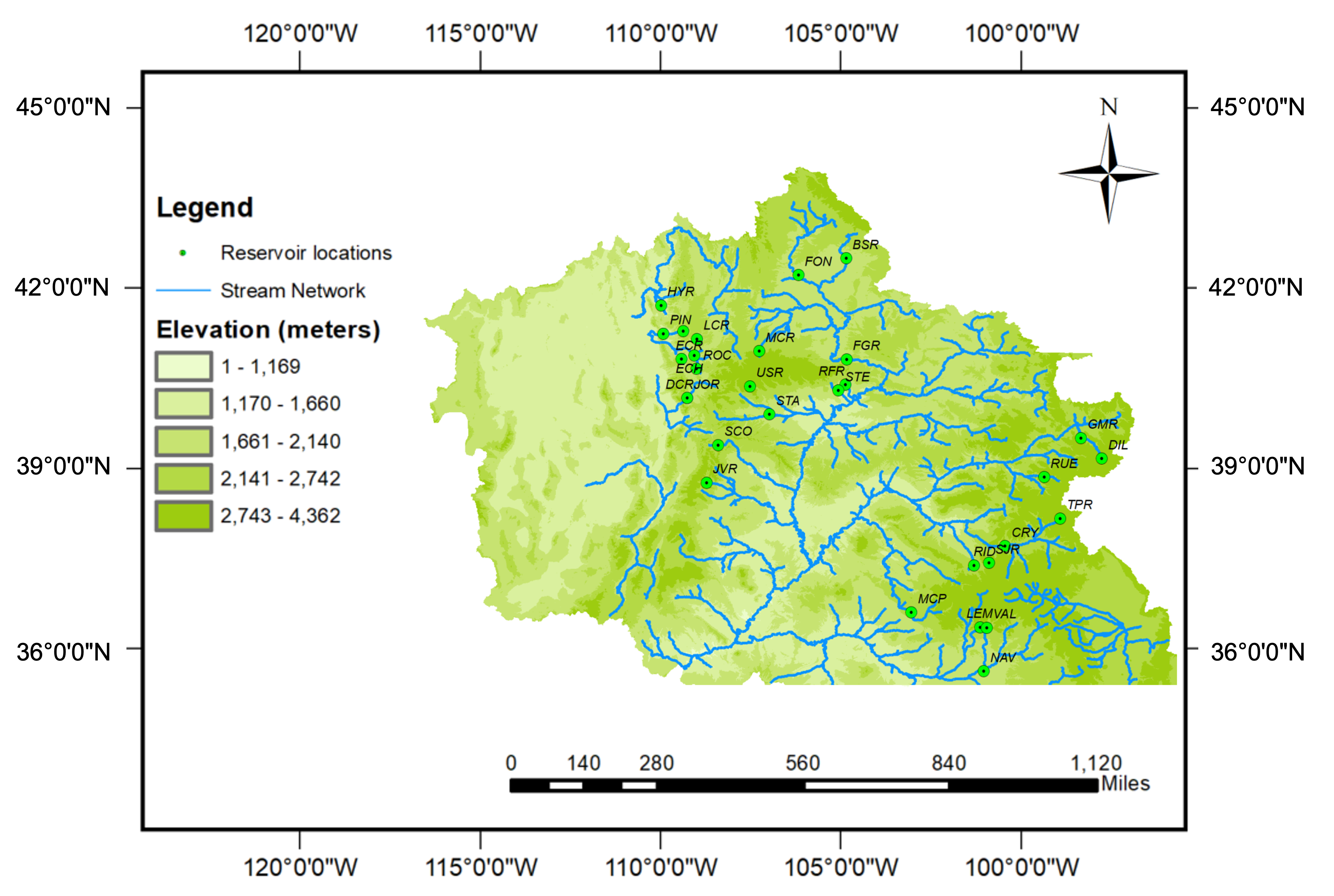}
    \caption{Geographical Position and River Network among 30 reservoirs in the Upper Colorado River Basin.}
    \label{fig:geo_stats}
\end{figure}

Each reservoir is treated as a distinct domain. We use the observational window of 1999–2011, when all 30 reservoirs have continuous records. 
MCR, JVR, and MCP contribute only 3 years of observations and are designated as the target domain set, while the remaining reservoirs form the source domain set.
These target reservoirs are chosen because they are located on isolated stream branches without shared inflow from neighboring reservoirs, which increases their distributional gap from the source domain and yields a more reliable test of domain generalization. 
Note that, source reservoirs can be further separated into training and validation subsets for model development, ensuring no target-domain information is used in training.
The geographical information is illustrated in Figure~\ref{fig:geo_stats}.

\subsection{Evaluation Metrics}
\label{ap:metric}
The model’s prediction accuracy is assessed using the Nash--Sutcliffe Efficiency (NSE) metric, which quantifies the consistency between the predicted reservoir inflow and its observed values, as in Eq. (\ref{eq:nse}),
\begin{equation}
\text{NSE}(\hat{y}_i, {y}_i) = 1 - \frac{\sum_{i=1}^{n} (\hat{y}_i - y_i)^2}{\sum_{i=1}^{n} (y_i - \bar{y}_i)^2}
\label{eq:nse}
\end{equation}
where $y_i$ and $\hat{y}_i$ are ground truths and predictions, $\bar{y}_i$ refers to the mean value of the observations, and $n$ represents the total number of observations. The range of the NSE is $(-\infty, 1]$, with 1 indicating perfect predictive performance.

\subsection{Baselines}
\label{ap:baselines}
We compare \model{} against three categories of baselines.
(1) The first category consists of two na\"ive baselines, including \textbf{Oracle}, trained directly on the target data, and \textbf{Base}, trained solely on the source data. This setting is commonly used for evaluation by recent studies~\citep{yang2023manydg, hu2025udoncare}.
(2) The second category comprises an intuitive solution \textbf{Few-shot}, trained on a split of the target data (train/test on target). 
(3) The last category consists of four conventional DG methods, which are still runnable within such fine-grained domain categorization:
\begin{enumerate}
    \item \textbf{CondAdv}~\cite{zhao2017learning}: This conditional adversarial network concatenates the predicted class-probability vector with the feature embedding, and feeds the combined representation into a domain discriminator trained adversarially to confuse domain classification. 
    \item \textbf{DANN}~\cite{ganin2016domain}: Domain-Adversarial Neural Network uses a gradient-reversal layer inserted between feature extractor and domain classifier. It flips the sign of gradient from domain classifier to encourage the feature extractor to produce domain-invariant representations, while maintaining accuracy on the source domain.
    \item \textbf{IRM}~\cite{li2022invariant}: Invariant Risk Minimization aims to find a representation such that the optimal classifier performs well across all source domains. It enforces a penalty on the squared gradient norm of the classifier’s risk with respect to the representation, promoting invariance across domain-specific risks.
    \item \textbf{MLDG}~\cite{li2018learning}: Meta-Learning for Domain Generalization simulates domain shift during mini-batch training by splitting source domains into “meta-train” and “meta-test” subsets, such that improvement on meta-train leads to improvement on meta-test.
\end{enumerate}

\begin{table*}[t]
\centering
\setlength{\tabcolsep}{1.8mm}
\begin{tabular}{l l c c c c c c c}
\toprule
\textbf{Method} & \textbf{Overall} & \textbf{Day 1} & \textbf{Day 2} & \textbf{Day 3} & \textbf{Day 4} & \textbf{Day 5} & \textbf{Day 6} & \textbf{Day 7} \\
\midrule
\multicolumn{9}{l}{\textit{Feature Encoder (Backbone) Only}} \\
\cmidrule(lr){1-9}
\textbf{Base} \texttt{(Lower Bound)}          & 78.29\textsubscript{(1.7)}   & 87.96\textsubscript{(1.2)} & 84.78\textsubscript{(2.3)} & 81.92\textsubscript{(1.0)} & 78.95\textsubscript{(1.9)} & 75.34\textsubscript{(2.7)} & 71.16\textsubscript{(1.4)} & 67.95\textsubscript{(2.0)} \\
\textbf{Few-shot}      & 80.08\textsubscript{(1.3)} & 89.08\textsubscript{(1.8)} & 86.46\textsubscript{(0.9)} & 82.28\textsubscript{(2.1)} & 80.19\textsubscript{(1.7)} & 77.32\textsubscript{(1.0)} & 74.20\textsubscript{(2.6)} & 71.03\textsubscript{(1.3)} \\
\textbf{Oracle} \texttt{(Upper Bound)}       & 83.93\textsubscript{(2.1)}   & 93.79\textsubscript{(2.3)} & 90.64\textsubscript{(1.4)} & 86.91\textsubscript{(2.7)} & 83.25\textsubscript{(0.7)} & 80.54\textsubscript{(1.9)} & 77.09\textsubscript{(2.5)} & 75.29\textsubscript{(0.9)} \\
\midrule
\multicolumn{9}{l}{\textit{Domain Generalization Baselines}} \\
\cmidrule(lr){1-9}
\textbf{DANN}~\cite{ganin2016domain}          & 78.89\textsubscript{(1.1)} & 88.63\textsubscript{(1.5)} & 85.12\textsubscript{(2.1)} & 82.34\textsubscript{(0.8)} & 79.50\textsubscript{(2.7)} & 75.79\textsubscript{(1.1)} & 71.92\textsubscript{(2.5)} & 68.94\textsubscript{(2.0)} \\
\textbf{MLDG}~\cite{li2018learning}          & 80.67\textsubscript{(2.2)} & 89.82\textsubscript{(2.0)} & 87.11\textsubscript{(1.7)} & 83.71\textsubscript{(2.4)} & 80.53\textsubscript{(1.2)} & 77.66\textsubscript{(0.9)} & 74.43\textsubscript{(2.5)} & 71.42\textsubscript{(1.8)} \\
\textbf{CondAdv}~\cite{zhao2017learning}       & 80.77\textsubscript{(1.9)} & 90.06\textsubscript{(1.0)} & 86.99\textsubscript{(2.6)} & 83.62\textsubscript{(1.4)} & 80.73\textsubscript{(2.4)} & 77.80\textsubscript{(1.2)} & 74.58\textsubscript{(1.9)} & 71.61\textsubscript{(2.1)} \\
\textbf{IRM}~\cite{li2022invariant}           & 78.50\textsubscript{(2.4)} & 88.15\textsubscript{(2.7)} & 84.98\textsubscript{(1.6)} & 82.12\textsubscript{(2.2)} & 79.13\textsubscript{(0.9)} & 75.52\textsubscript{(1.6)} & 71.38\textsubscript{(2.4)} & 68.20\textsubscript{(1.0)} \\
\midrule
\multicolumn{9}{l}{\textit{Our Method}} \\
\cmidrule(lr){1-9}
\textbf{\model{}}      & \textbf{82.90\textsubscript{(1.4)}} & \textbf{92.92\textsubscript{(1.8)}} & \textbf{89.60\textsubscript{(0.8)}} & \textbf{86.03\textsubscript{(2.2)}} & \textbf{82.26\textsubscript{(1.3)}} & \textbf{79.26\textsubscript{(2.7)}} & \textbf{76.24\textsubscript{(0.9)}} & \textbf{73.96\textsubscript{(1.5)}} \\
\quad - w/o Contrastive Loss & 80.30\textsubscript{(1.6)} & 89.41\textsubscript{(1.4)} & 86.84\textsubscript{(1.2)} & 83.35\textsubscript{(1.8)} & 80.09\textsubscript{(1.0)} & 77.37\textsubscript{(1.7)} & 74.10\textsubscript{(1.1)} & 70.94\textsubscript{(1.9)} \\
\quad - w/o Adversarial Loss & 79.09\textsubscript{(1.0)} & 88.81\textsubscript{(1.6)} & 85.39\textsubscript{(1.5)} & 82.49\textsubscript{(0.9)} & 79.74\textsubscript{(1.8)} & 75.98\textsubscript{(1.2)} & 72.13\textsubscript{(1.7)} & 69.06\textsubscript{(1.1)} \\
\quad - w/o FiLM Adaptor & 81.39\textsubscript{(1.5)} & 90.40\textsubscript{(1.9)} & 88.02\textsubscript{(1.1)} & 84.37\textsubscript{(1.6)} & 81.32\textsubscript{(0.8)} & 78.30\textsubscript{(1.4)} & 75.31\textsubscript{(1.2)} & 71.99\textsubscript{(1.0)} \\
\quad - w/ Spatial Shuffle & 80.63\textsubscript{(1.8)} & 89.66\textsubscript{(1.2)} & 87.34\textsubscript{(1.6)} & 82.85\textsubscript{(1.4)} & 80.82\textsubscript{(1.1)} & 77.22\textsubscript{(1.7)} & 74.37\textsubscript{(0.9)} & 72.10\textsubscript{(1.3)} \\
\bottomrule
\end{tabular}
\caption{Overall and daily NSE scores from Day 1 to 7 across three target reservoirs (MCR, JVR, MCP) for all baselines. 
We report the average performance (\%) and the standard deviation (in bracket) of each model over 5 runs.}
\label{tab:main-res}
\end{table*}

\subsection{Implementation Details}
We randomly initialize all embeddings and model parameters.
The hidden dimensions of $f_\phi(\cdot)$, $d_\theta(\cdot)$, $m_\delta(\cdot)$, and $p_\omega(\cdot)$ are set to 64, 32, 64, and 64, respectively. 
The feature extractor $f_\phi(\cdot)$ adopts an Encoder–Decoder LSTM structure with two layers and a hidden size of 64, consistent with the configuration in~\citep{fan2023advancing}. 
$\mathbf{X}_i \in \mathbb{R}^{30 \times 3}$ contains temperature, precipitation, and inflow over the past 30 days, while the spatial vector $\mathbf{s}_i \in \mathbb{R}^3$ represents latitude, longitude, and elevation. 
We apply dropout rate 0.1 across layers, and we train for 100 epochs using the Adam optimizer with an initial learning rate of $1\times10^{-3}$, decayed by a factor of 0.5 through a ReduceLROnPlateau scheduler (patience = 10). 
The batch size is 32, and $L_{\text{sup}}$ is mean squared error. 
The loss weights are fixed as $\lambda_{\text{sup}}=1.0$ and $\lambda_{\text{adv}}=0.1$, with a warm-up period of 10 epochs before adversarial training begins. 
Gradient clipping with a maximum norm of 1.0 is applied to stabilize optimization.

Training proceeds in two stages: the first 10 epochs use $L_{\text{con}}$ and $L_{\text{sup}}$ to learn invariant representations, while subsequent epochs incorporate $L_{\text{adv}}$ with feature modulation. 
During inference, only $f_\phi(\cdot)$, $m_\delta(\cdot)$, and $p_\omega(\cdot)$ are activated, and the model outputs 7-day ahead inflow predictions for all target reservoirs. 
All experiments are implemented in Python~3.10.17, PyTorch~2.5.1, and PyG~2.6 with ROCm~6.2.4 on a server node equipped with three AMD EPYC 64-core CPUs, 512~GB RAM, and eight AMD MI250X GCDs.

\subsection{Experiment Results}
\subsubsection{Overall Performances.}
Following the baseline settings, \textbf{Base} is trained using data from 27 source reservoirs and directly evaluated on 3 target reservoirs, which represents the basic performance (i.e. lower bound) without domain adaptation. 
Similarly, \textbf{Oracle} is trained and evaluated on target reservoirs, assuming full access to target data; this setting is unrealistic in practice but provides the upper bound. 
\textbf{Few-shot} divides target reservoirs into training and testing splits, where the first two years are used for training and the most recent year for testing, reflecting a practical limited-data setting.
Table~\ref{tab:main-res} shows a clear and consistent gap between Oracle and Base, which confirms domain shift across target reservoirs.
Such gap also appears on each forecast day, which further justifies the DG setting. 
Few-shot also demonstrates the improvement over Base, ensuring the effectiveness of leveraging limited in-domain samples to mitigate domain shifts. 
However, \model{}, MLDG, and CondAdv surpass Few-shot without relying on full target supervision, which indicates that DG solutions might be more competitive in hydrological settings.
Moreover, \model{} demonstrates superior NSE scores across forecasting days, close to Oracle ($-1.03\%$), over all DG baselines.

\begin{figure*}[t]
    \centering
    \includegraphics[width=\linewidth]{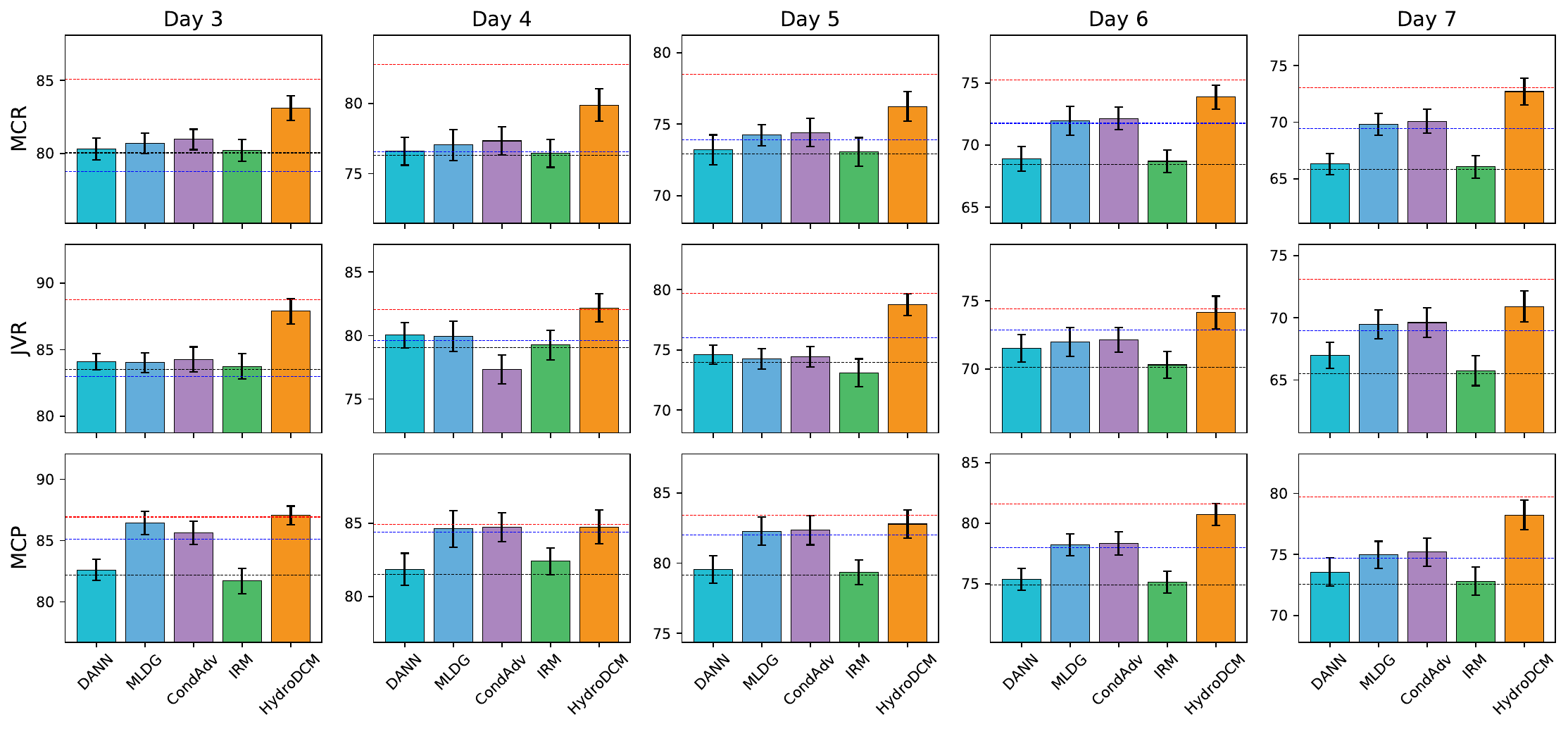}
    \caption{\textbf{Comparison of model performance on daily NSE scores across three reservoirs (MCR, JVR, and MCP) from Day 3 to Day 7.}
    Each subplot shows the NSE (\%) for \textbf{DANN}, \textbf{MLDG}, \textbf{CondAdv}, \textbf{IRM}, and \textbf{HydroDCM} on a specific reservoir–day combination. 
    Error bars indicate standard deviations over five independent runs. 
    The horizontal black, blue, and red dash lines represent the \textbf{Base}, \textbf{Few-shot}, and \textbf{Oracle} standards for observing enhanced robustness, respectively.}
    \label{fig:individual_main}
\end{figure*}

\subsubsection{Individual Performances.}
Figure~\ref{fig:individual_main} shows individual results on target reservoirs from Day~3 to Day~7. 
We focus on these horizons because prediction accuracy typically declines as the lead time increases, and thus the later days provide a more stringent test of model generalization. 
Across all three reservoirs, \model{} consistently achieves the highest NSE scores among other DG baselines. 
For instance, at Day~7, \model{} improves NSE by roughly $2.5\%\!\sim\!3.0\%$ over MLDG and CondAdv, and more than $4\%$ over IRM and DANN on average. 
Moreover, the rate of performance degradation with increasing forecast horizon is significantly lower for \model{}, suggesting that its domain-conditioned modulation effectively preserves temporal stability as predictive uncertainty accumulates. Overall, these findings demonstrate that \model{} achieves superior robustness and generalization, even as forecasting uncertainty grows with longer lead times.

Compared with the three lines, \model{} also exhibits substantial advantages. 
Relative to Base, \model{} yields an average improvement of $3\%\!\sim\!5\%$. 
Compared with the Few-shot setting, \model{} still achieves $1\%\!\sim\!3\%$ higher NSE on later forecast days, and in some cases approaches or even matches the Oracle upper bound trained directly on target data. 
The advantage arises from the data scarcity in target reservoirs, which offer only three-year observations in this work. 
Such limited samples prevent Few-shot models from learning stable temporal patterns, whereas \model{} compensates for this deficiency by transferring cross-reservoir information through modulation.
Overall, these results confirm that \model{} not only maintains higher accuracy at short horizons but also exhibits markedly lower degradation over time, highlighting its superior robustness and generalization capability in multi-domain hydrological forecasting.

\subsubsection{Ablation Study.}
We evaluate four variants to disentangle individual contributions of each component in our method:
\begin{itemize}
    \item \textbf{w/o Contrastive Loss}: remove the contrastive objective by setting its weight $\lambda_{\text{con}}=0$, effectively dropping the module that enforces inter-domain discriminability in the representation space.
    \item \textbf{w/o Adversarial Loss}: remove the domain-adversarial objective by setting $\lambda_{\text{adv}}=0$, disabling the alignment pressure that suppresses domain covariates.
    \item \textbf{w/o FiLM Adaptor}: discard the modulation of the invariant feature $\mathbf{z}$, i.e., no FiLM-based spatial conditioning is applied during inference.
    \item \textbf{w/ Spatial Shuffle}: corrupt the spatial side information by injecting Gaussian noise into the metadata and randomly shuffling them across sites, preventing the model from leveraging informative spatial cues.
\end{itemize}
In Table~\ref{tab:main-res}, we can observe that: (1) removing the adversarial loss causes the largest degradation, since \model{} cannot strip domain covariates from input features without adversarial alignment; 
(2) deprecating the contrastive loss or corrupting spatial information both reduce accuracy, indicating that clear inter-domain separation is needed to learn label-relevant structure and that valid spatial metadata provide useful signal; 
(3) skipping the FiLM adaptor leads to a smaller but consistent drop across days, showing that injecting faithful spatial context to modulate $\mathbf{z}$ further improves hydrological prediction quality in practice. 
Overall, the full \model{} that combines adversarial alignment, contrastive separation, and FiLM-based spatial conditioning achieves the most stable and accurate NSE from Day~1 to 7.

\section{Conclusion}
We present \model{}, a novel domain generalization framework tailored for cross-reservoir inflow prediction in hydrological systems.
It bridges the gap between domain invariance and reservoir-specific adaptability by incorporating spatial metadata as pseudo-domain guidance. 
Extensive experiment results on real reservoirs in the Upper Colorado River Basin demonstrate the superiority of \model{} over existing baselines across all forecast horizons. 
Ablation results further confirm the contribution of each module and the complementary role of geographic attributes in achieving robust generalization.
Overall, this work constitutes the first exploration of domain generalization for hydrological forecasting across many domains, providing practical insights into scalable learning under spatial heterogeneity. 


\section*{Acknowledgments}
This research was supported by the Seedling Project, funded by the U.S. Department of Energy (DOE) Water Power Technologies Office, and by Dan Lu’s Early Career Project, funded by the DOE Biological and Environmental Research Office. Additional support was provided through collaboration between the U.S. Air Force Life Cycle Management Center (LCMC) and Oak Ridge National Laboratory (ORNL). High-performance computing resources were provided by the Frontier system at the Oak Ridge Leadership Computing Facility, a DOE Office of Science User Facility.
This work was supported in part by the U.S. National Science Foundation under grants 2047843 and 2437621.

\bibliography{main}

\appendix
\section*{Appendix}
\section{Additional Copyright Notice}
\label{sec:gov-notice}
This manuscript has been authored by UT-Battelle LLC, under contract DE-AC05-00OR22725 with the US Department of Energy (DOE). The US government retains and the publisher, by accepting the article for publication, acknowledges that the US government retains a nonexclusive, paid-up, irrevocable, worldwide license to publish or reproduce the published form of this manuscript, or allow others to do so, for US government purposes. DOE will provide public access to these results of federally sponsored research in accordance with the DOE Public Access Plan (http://energy.gov/downloads/doe-public-access-plan).

\section{Further Discussion}

While \model{} demonstrates strong generalization performance across hydrological domains, several promising extensions remain open for future exploration:

\textbf{1) Enriching Domain Descriptors.}
Currently, the domain conditioning in \model{} relies mainly on spatial attributes. Extending the metadata space to incorporate hydrological, climatic, and anthropogenic descriptors (e.g. land use, soil permeability, and climate indices) may help capture more diverse forms of domain heterogeneity. 
Future work may involve fusing these heterogeneous descriptors through attention-based or graph-based metadata encoders to enhance model adaptability while maintaining interpretability with respect to underlying hydrological processes.

\textbf{2) About Adversarial–Contrastive Formulation.}
The proposed formulation balances domain invariance and discriminability in practice. Nevertheless, a more rigorous theoretical understanding is warranted. Future research may investigate its sensitivity to factors such as the contrastive temperature parameter $\tau$, embedding dimensionality, and the weighting of loss components. Analyzing the method within an information-theoretic or causal representation framework could reveal how pseudo-domain embeddings interact with adversarial alignment, providing insights for adaptive tuning strategies across varying hydrological regimes.

\textbf{3) Expanding the Evaluation Landscape.}
Future benchmarks may include more recent domain generalization and transfer-learning methods. In addition, \textit{Oracle} baseline still offers only a simplified upper bound trained on target data. Introducing transfer learning or fine-tuning baselines would provide a more realistic performance reference for practical deployment scenarios.

\textbf{4) Toward Adaptable Generalization.}
One potential direction is to integrate \model{} with physical constraints and multi-modal hydrological data. 
Combining domain generalization with physics-informed modeling or remote-sensing information could enhance both robustness and interpretability. 
Such integration would help bridge purely data-driven generalization and process-based hydrological reasoning, enabling more reliable predictions under nonstationary or extreme climate conditions.

In summary, extending \model{} along these directions will further advance its potential as a foundation framework for cross-domain hydrological forecasting.

\end{document}